\def\year{2020}\relax
\documentclass[letterpaper]{article} %
\usepackage{aaai20}  %
\usepackage{times}  %
\usepackage{helvet} %
\usepackage{courier}  %
\usepackage[hyphens]{url}  %
\usepackage{graphicx} %
\usepackage{graphicx}  %
\usepackage{algorithm}
\usepackage[noend]{algpseudocode}
\usepackage{subcaption}
\usepackage{caption}
\usepackage{booktabs} 
\usepackage{stmaryrd} 
\usepackage{xcolor}
\usepackage{mathtools}

\usepackage{color} %

\title{Time-based Dynamic Controllability of Disjunctive Temporal Networks with Uncertainty: A Tree Search Approach with Graph Neural Network Guidance}

\author{
    \parbox{\linewidth}{\centering
        Kevin Osanlou\textsuperscript{1,2,3,4},
        Jeremy Frank\textsuperscript{1},
        J. Benton\textsuperscript{1},
        Andrei Bursuc\textsuperscript{5},
        Christophe Guettier\textsuperscript{2},
        Eric Jacopin\textsuperscript{6} and
        Tristan Cazenave\textsuperscript{3}
    } 
    \\
    {\fontsize{11}{13}\selectfont$^1$ NASA Ames Research Center \qquad  $^2$ Safran Electronics \& Defense \qquad $^3$LAMSADE, Paris-Dauphine}\\
    {\fontsize{11}{13}\selectfont $^4$ Universities Space Research Association \qquad $^5$valeo.ai \qquad $^6$CREC Saint-Cyr Coetquidan}\\

    \tt\small{\{kevin.osanlou, jeremy.d.frank, j.benton\}@nasa.gov}\\ 
    \tt\small{\{kevin.osanlou, christophe.guettier\}@safrangroup.com} \\
	\tt\small{andrei.bursuc@valeo.com} \\ \tt\small{eric.jacopin@st-cyr.terre-net.defense.gouv.fr} \\
	\tt\small{tristan.cazenave@lamsade.dauphine.fr}
	
}

\begin{document}

\maketitle

\def\fromjeremy#1{{\color{blue}\small{\bf JF:} {\em #1}}}
\def\fromkevin#1{{\color{red}\small{\bf KO:} {\em #1}}}
\def\fromJ#1{{\color{yellow}\small{\bf JB:} {\em #1}}}
\def\fromEric#1{{\color{pink}\small{\bf EJ:} {\em #1}}}
\def\fromTristan#1{{\color{purple}\small{\bf TC:} {\em #1}}}
\def\fromAndrei#1{{\color{brown}\small{\bf AB:} {\em #1}}}
\def\fromChristophe#1{{\color{green}\small{\bf CG:} {\em #1}}}

\newcommand{\ab}[1]{\textcolor{orange}{#1}}
\newcommand{\abc}[1]{\textcolor{orange}{[AB: #1]}}

\def\authassign#1{{\color{orange}\small{\bf Contributor(s):} {\em #1}}}
\def\dor{\textit{d-OR }}
\def\wor{\textit{w-OR }}
\def\and{\textit{AND }}
\def\wait{\textit{WAIT }}
\def\etal{\emph{et al.}}
\def\ie{\emph{i.e. }}
\def\eg{\emph{e.g. }}
\def\True{\emph{True }}
\def\true{\emph{true }}
\def\False{\emph{False }}
\def\false{\emph{false }}

\begin{abstract}
Scheduling in the presence of uncertainty is an area of interest in artificial intelligence due to the large number of applications. %
We study the problem of dynamic controllability (DC) of disjunctive temporal networks with uncertainty (DTNU),
which seeks a strategy to satisfy all constraints in response to uncontrollable action durations.
We introduce a more restricted, stronger form of controllability than DC for DTNUs, time-based dynamic controllability (TDC), and present a tree search approach to determine whether or not a DTNU
is TDC. Moreover, we leverage the learning capability of a message passing neural network (MPNN) 
as a heuristic for tree search guidance. Finally, we conduct experiments for which the tree search shows superior results to state-of-the-art timed-game automata (TGA) based approaches. We observe that using an MPNN for tree search guidance leads to a significant increase in solving performance and scalability to harder DTNU problems.

\end{abstract}

\section{Introduction}

\begin{figure}[t!]
\renewcommand{\captionfont}{\small}
\centering
\includegraphics[scale=0.57]{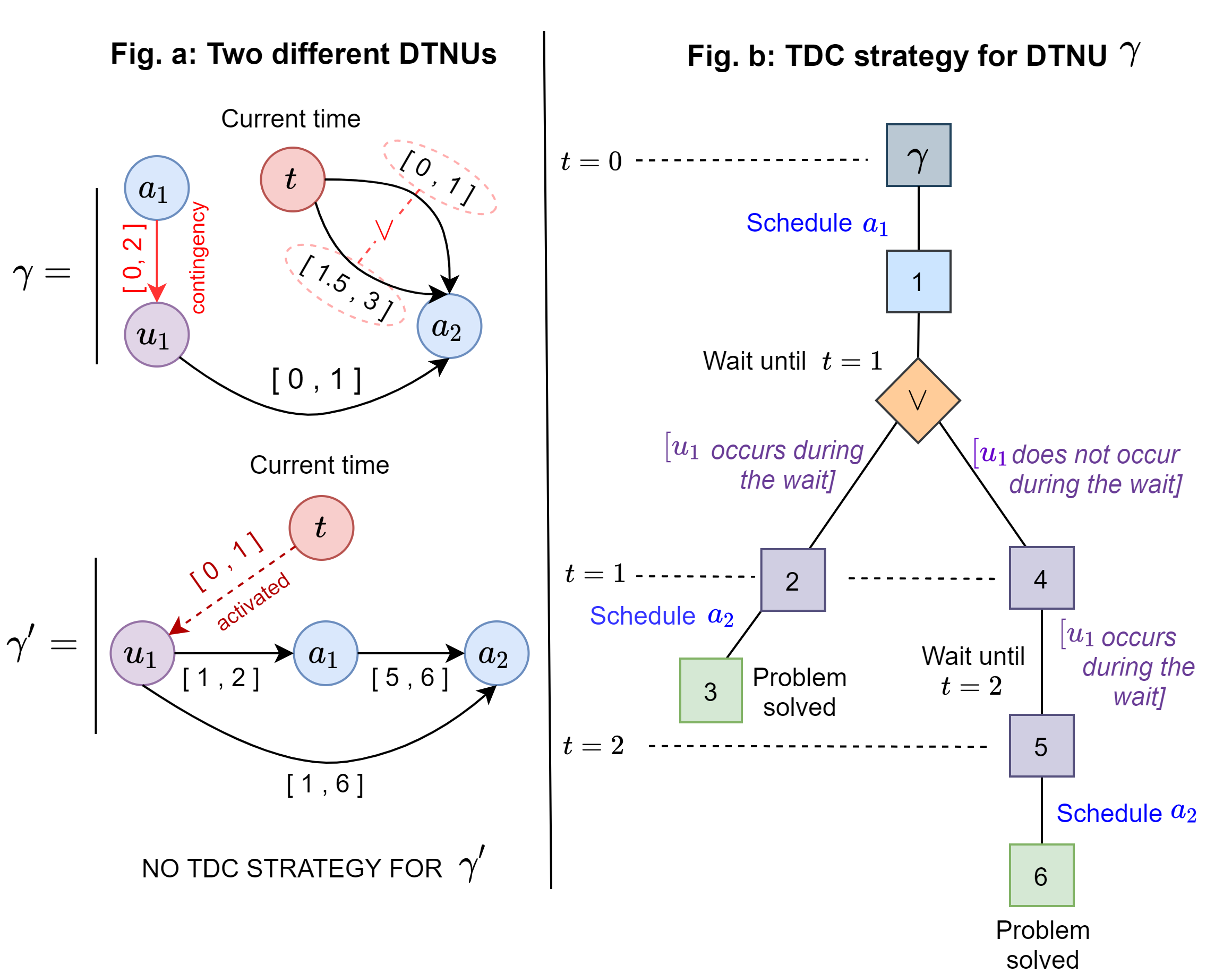}
\caption{\textbf{Two example DTNUs $\gamma$ and $\gamma'$.} In both examples, timepoints $a_1$ and $a_2$ are controllable; $u_1$ is uncontrollable. Black arrows and their intervals (valued in ${\rm I\!R}$) represent time constraints between timepoints; the light red arrow and its interval contingency links. The dashed dark red arrow in $\gamma'$ implies $u_1$ has already been activated and will occur in the specified interval. A TDC strategy is displayed for $\gamma$: the root node of the strategy is $\gamma$ while other nodes are sub-DTNUs except the $\lor$ node which lists transitional possibilities. DTNU $\gamma'$, on the other hand, is an example of a DTNU which is DC but not TDC.}
\label{fig:easy-example}
\vspace{-0.4cm}
\end{figure}

Temporal Networks (TN) are a common formalism to represent and reason about temporal constraints over a set of time points (e.g. start/end of activities in a scheduling problem). The Simple Temporal Networks with Uncertainty (STNUs) \cite{kn:Ts} \cite{kn:ViFa}
explicitly incorporate qualitative uncertainty into temporal networks.  
Considerable work has resulted in algorithms to determine whether or not all timepoints can be scheduled, either up-front or reactively, in order to account for uncertainty (e.g. \cite{kn:MoMu2}, \cite{kn:Mofast}). 
In particular, an STNU is {\em dynamically controllable} (DC) if there is a reactive
strategy in which controllable timepoints can be executed
either at a specific time, or after observing the occurrence of an uncontrollable timepoint.
Cimatti et al. \cite{cimatti2016dynamic} investigate the problem of DC for 
Disjunctive Temporal Networks with Uncertainty
(DTNUs), which generalize STNUs.  
Figure~\ref{fig:easy-example}a
shows an example of two DTNUs $\gamma$ and $\gamma'$ on the left side; $a_i$ are controllable timepoints, $u_j$ are uncontrollable timepoints. Timepoints are variables which can take on any value in ${\rm I\!R}$. Constraints between timepoints characterize both a minimum and maximum time distance separating them, likewise valued in ${\rm I\!R}$. The key difference between STNUs and DTNUs
lies in the {\em disjunctions} that yields more choice points for consistent scheduling, especially reactively.

The complexity of DC checking for DTNUs is $PSPACE$-complete
\cite{kn:BhWi}, making this a very challenging problem.
The difficulty in proving or disproving DC arises from the need to check all possible combinations of disjuncts in order to handle all possible occurrence outcomes of the uncontrollable timepoints. The best previous approaches for this problem use timed-game automata (TGAs) and Satisfiability Modulo Theories (SMTs), described in \cite{cimatti2016dynamic}.

Recently, applications such as image classification have benefited from learning techniques such as Convolutional Neural Networks (CNNs) \cite{krizhevsky_2012}.
A new emerging trend of neural networks, graph-based neural networks (GNNs), have been proposed as an extension of CNNs to graph-structured data; recent variants based on spectral graph theory include  \cite{defferrard_2016}, \cite{YujiaLi2016}, \cite{kipf_2017}. These GNNs take advantage of relational properties between nodes for classification, but do not take into account potential edge weights. In newer approaches, Message Passing Neural Networks (MPNNs) with architectures such as in \cite{battaglia2016interaction}, \cite{gilmer2017neural} and \cite{kipf2018neural} use embeddings comprising edge weights within each computational layer. We focus our interest on these architecture types as DTNUs can be formalized as graphs with edge distances representing time constraints.

In this work, we study DC checking of DTNUs as a %
search problem, express %
states as graphs, and use MPNNs to learn heuristics based on previously solved DTNUs to guide search.
The key contributions of our approach are the following. \textbf{(1)} We introduce a time-based form of dynamic controllability (TDC) and a tree search approach to identify TDC strategies. We informally show that TDC implies DC, but the opposite is not generally true.
\textbf{(2)} We define a relevant way of using an MPNN architecture for handling DTNU scheduling problems and use it as heuristic for guidance in the tree search. Moreover, we define a  self-supervised training scheme to train the MPNN based on solving randomly generated DTNUs with short timeouts to limit search duration.
\textbf{(3)} We introduce constraint propagation rules which enable us to enforce time domain restrictions for variables in order to ensure soundness of strategies found. We carry out experiments which show the tree search algorithm offers improved scalability over the best previous DC-solving approach evaluated in \cite{cimatti2016dynamic}, PYDC-SMT. Moreover, we show the tree search is improved upon significantly by the learned MPNN heuristic on harder DTNUs.

\section{Time-based Dynamic Controllability}
\label{tdc}

A DC {\em strategy} for a DTNU either executes controllable timepoints at a specific time, or reacts to the occurrence of an uncontrollable timepoint. We present our TDC formalism here.
A TDC strategy executes controllable timepoints at specific times under the assumption that some uncontrollable timepoints may occur or not in a given time interval. Each interval in a TDC strategy can have an arbitrary duration.  Thus, controllable timepoints are usually executed at the end of the same interval, regardless of if and when uncontrollable timepoints occur inside the interval. Within a given interval, TDC also leaves open the choice to execute a controllable timepoint at the same time as the occurrence of an uncontrollable timepoint, which we call reactive execution.

Nonetheless, TDC is less flexible than a DC strategy which can wait for an uncontrollable timepoint to occur before making a new decision. It does not allow, for example, {\em delayed} reactive execution of the controllable timepoint.  TDC is a subset of DC, and a stronger form on controllability: TDC implies DC. As described below, representing reactive DC strategies in TDC may require the tree to become arbitrarily large, so DC does not imply TDC. 
DTNU $\gamma'$ in figure~\ref{fig:easy-example}a shows an example of an STNU which is not TDC but DC. In this example, uncontrollable timepoint $u_1$ is activated, \ie the controllable timepoint associated to $u_1$ in the contingency links has been executed. Moreover, it is known that $u_1$ occurs between $t$ and $t + 1$, where $t$ is the current time. The interval $[t,t+1]$ is referred to as the \emph{activation time interval} for  $u_1$. Controllable timepoint $a_1$ must be executed at least 1 time unit after $u_1$, and controllable timepoint $a_2$ at least 5 time units after $a_1$. However, controllable timepoint $a_2$ cannot be executed later than 6 time units after $u_1$. A valid DC strategy waits for $u_1$ to occur, then schedules $a_1$ exactly 1 time unit later, and $a_2$ 5 time units after $a_1$. However, for any TDC strategy, there is no wait duration small enough while waiting for $u_1$ to happen that does not violate these constraints. There will always be some strictly positive lapse of time between the moment $u_1$ occurs and the end of the wait. The exact execution time of $u_1$ during the wait is unknown: a TDC strategy therefore assumes $u_1$ happened at the end of the wait when trying to schedule $a_1$ at the earliest.
Therefore, the earliest time $a_1$ 
can be scheduled in a TDC strategy is 1 time unit after the end of the wait, which is too late.

\section{Tree Search Preliminaries}
\label{tree-search-prelim}

We introduce here the tree search algorithm. Intuitively, the approach discretizes uncontrollable durations, \ie durations when one or several uncontrollable timepoints can occur, into multiple reduced intervals. These reduced intervals are then used to account for possible outcomes of uncontrollable timepoints and adapt the scheduling strategy accordingly. 

The root of the search tree is the DTNU, and other tree nodes are either sub-DTNUs of the DTNU or logical nodes (\textit{OR, AND}) which represent the presence or lack of control over transition into children nodes. At a given DTNU tree node, decisions such as executing a controllable timepoint or waiting for a period of time 
develop children DTNU nodes for which these decisions are propagated to constraints. The TDC controllability of a \textit{leaf} DTNU, \ie a sub-DTNU for which all controllable timepoints have been executed and uncontrollable timepoints are assumed to have occurred in specific intervals, indicates
whether or not this sub-DTNU has been solved at the end of the scheduling process. We also refer to the TDC controllability of a DTNU node in the search tree as its \textit{truth attribute}. Lastly, the search logically combines TDC controllability of children DTNUs to determine TDC controllability for parent nodes. We give a simple example of a TDC strategy for a DTNU $\gamma$ in figure~\ref{fig:easy-example}.

\begin{figure}[tb]
\renewcommand{\captionfont}{\small}
\centering
\includegraphics[scale=0.68]{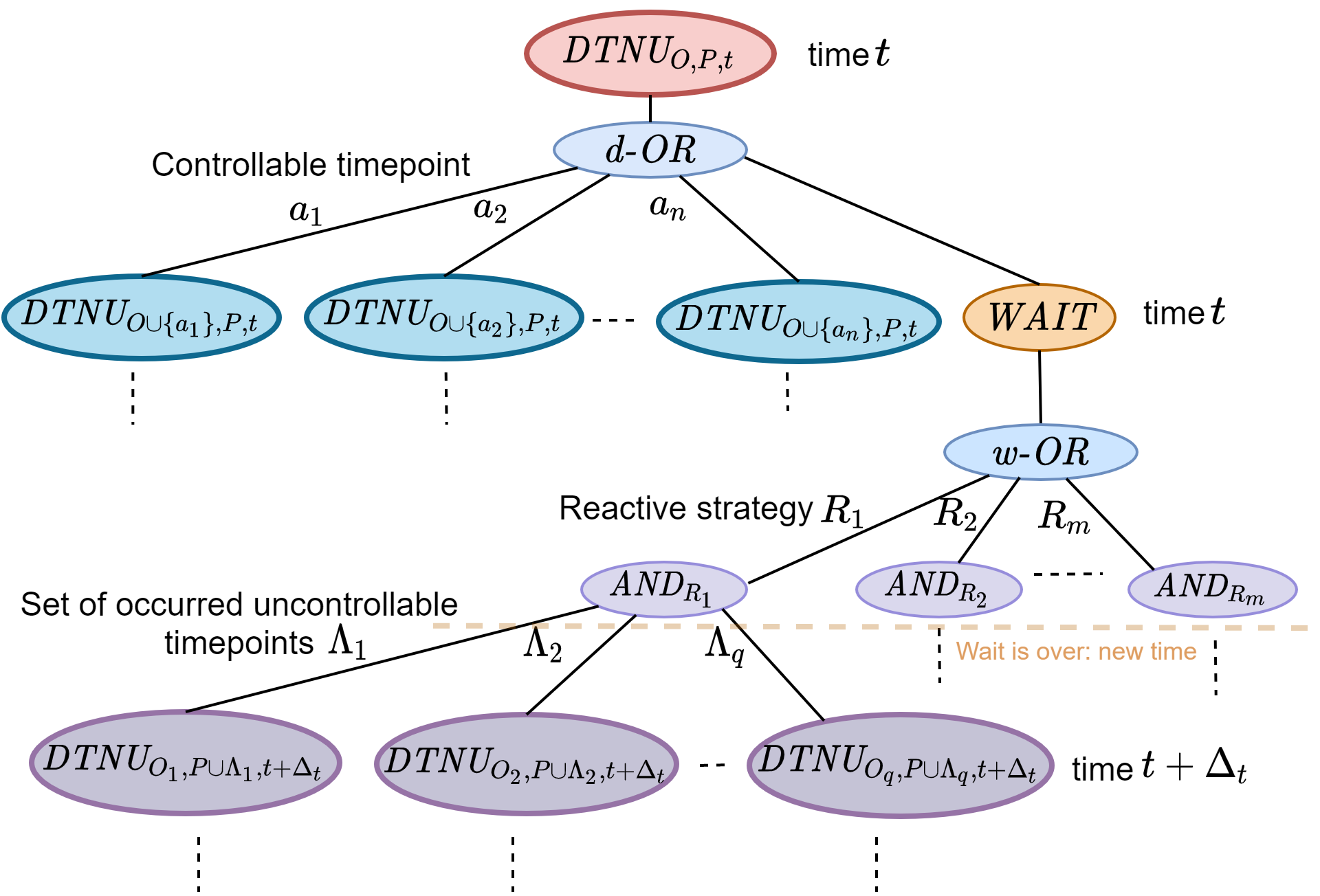}
\caption{\textbf{Basic structure of the search tree describing how a DTNU node \boldmath$DTNU_{O, P, t}$ is developed.} 
$DTNU_{O, P, t}$ (placed at the root of the tree) refers to a DTNU where $O$ is the set of controllable timepoints that have already been executed, $P$ the set of uncontrollable timepoints that have occurred, and $t$ the time. Each branch $a_i$ refers to a controllable timepoint $a_i$, $R_i$ to a reactive strategy during the wait, and $\Lambda_i$ to a combination of uncontrollable timepoints which can occur during the wait.
}
\label{fig:ts-structure-fig}
\vspace{-0.5 cm}
\end{figure}

Let $\Gamma = \{A,U,C,L\}$ be a DTNU. $A$ is the list of controllable timepoints, $U$ the list of uncontrollable timepoints, $C$ the list of constraints and $L$ the list of contingency links. The root node of the search tree built by the algorithm is $\Gamma$. There are four different types of nodes in the tree. Furthermore, each node possesses a \textit{truth} attribute, as explained in \S \ref{truthvalue}, which is initialized to \textit{unknown} and can be set to either \true or \false. 
The different types of tree nodes are listed below and shown in figure \ref{fig:ts-structure-fig}.

\paragraph{\textit{DTNU} nodes.} Any DTNU node other than the original problem $\Gamma$ corresponds to a sub-problem of $\Gamma$ at a given point in time $t$, for which some controllable timepoints may have already been scheduled in upper branches of the tree, some amount of time may have passed and some uncontrollable timepoints may have occurred. A DTNU node is made of the same timepoints $A$ and $U$, constraints $C$ and contingency links $L$ as the original DTNU $\Gamma$. It also carries a schedule memory $S$ of what exact time, or during what time interval, scheduled timepoints were executed during previous decisions in the tree. Lastly, the node also keeps track of the activation time intervals of activated uncontrollable timepoints $B$. The schedule memory $S$ is used to create an updated list of constraints $C'$ resulting from the propagation of the execution time or execution time interval of timepoints in constraints $C$ as described in \S \ref{tightbounds}. A non-terminal DTNU node, \ie a DTNU node for which all timepoints have not been scheduled, has exactly one child node: a \dor node.

\vspace{-0.4cm}
    
\paragraph{\textit{OR} nodes.} When a choice can be made at time $t$, this transition control is represented by an \textit{OR} node.
We distinguish two types of such nodes, \dor and \wor. For \dor nodes, the first type of choice available is which controllable timepoint $a_i$ to execute. This leads to a DTNU node.
The other type of choice is to wait a period of time (\S\ref{wait}) which leads to a \wait node.
\wor nodes can be used for \textit{reactive wait strategies}, \ie to stipulate  that some controllable timepoints will be scheduled reactively during waits (\S \ref{instant-scheduling}). 
The parent of a \wor node is therefore a \wait node and its children are \and nodes, described below.

\vspace{-0.4cm}

\paragraph{\wait nodes.} These nodes are used after a decision to wait a certain period of time $\Delta_t$. The parent of a \wait node is a \dor node. A \wait node has exactly one child: a \wor node, which has the purpose of exploring different reactive wait strategies. The uncertainty management related to uncontrollable timepoints is handled by \and nodes. %

\vspace{-0.4cm}
    
\paragraph{\and nodes.} %
This type of node represents a lack of transition control over children nodes. It is used after a wait decision is taken and a reactive wait strategy is decided, represented consecutively by a \wait and \wor node. Each child node of the \and node is a DTNU node at time $t + \Delta_t$, where $t$ is the time before the wait and $\Delta_t$ the wait duration. Furthermore, each child node represents an outcome of how uncontrollable timepoints may unfold, and is built from the set of {\em activated} uncontrollable timepoints (uncontrollable timepoints that have been started by the execution of their controllable timepoint) whose occurrence time interval overlaps the wait.  If there are $l$ activated uncontrollable timepoints, then there are at most $2^l$ \and node children, representing each element of the power set of activated uncontrollable timepoints (\S \ref{wait}).

Figure \ref{fig:ts-structure-fig} illustrates how a sub-problem of $\Gamma$, referred to as $DTNU_{O, P, t}$, is developed. Here, $O \subset A$ is the set of controllable timepoints 
that have already been executed, $P \subset U$ the set of uncontrollable timepoints which have occurred, and $t$ the time. This root node transitions into a \dor node. The \dor node in turn is developed into several children nodes $DTNU_{O \cup \{a_i\}, P,t}$ and a \wait node. Each node $DTNU_{O \cup \{a_i\}, P,t}$ corresponds to a sub-problem which is obtained from the execution of controllable timepoint $a_i$ at time $t$. The \wait node refers to the process of waiting a given period of time, $\Delta_t$ in the figure, before making the next decision. The \wait node leads directly to a \wor node which lists different wait strategies $R_i$. 
If there are $l$ activated uncontrollable timepoints, there are $2^l$ subsets of uncontrollable timepoints $\Lambda_i$ that could occur. Each $\textit{AND}_{R_j}$ node has one sub-problem DTNU for each $\Lambda_i$.

Each sub-problem $DTNU_{O_i, P \cup \Lambda_i, t + \Delta_t}$ of the node $\textit{AND}_{R_j}$ is a DTNU at time $t+\Delta_t$ for which all uncontrollable timepoints in $\Lambda_i$ are assumed to have happened during the wait period, \ie in the time interval $[t, t + \Delta_t]$. Additionally, some controllable timepoints may have been reactively executed during the wait and may now be included in the set of scheduled controllable timepoints $O_i$. Otherwise, $O_i = O$. 

Two types of leaf nodes exist in the tree. The first type is a node $DTNU_{A, U, t}$ for which all controllable timepoints $a_i \in A$ have been scheduled and all uncontrollable timepoints  $u_i \in U$ have occurred. The second type is a node $DTNU_{A \setminus A', U, t}$ for which all uncontrollable timepoints $u_i \in U$ have occurred, but some controllable timepoints  $a_i \in A'$ have not been executed. The constraint satisfiability test of the former type of leaf node is straightforward: all execution times of all timepoints are propagated to constraints in the same fashion as in \S \ref{tightbounds}. The leaf node's truth attribute is set to \true if all constraints are satisfied, \false otherwise. For the latter type, we propagate the execution times of all uncontrollable timepoints as well as all scheduled controllable timepoints in the same way, and obtain an updated set of constraint $C'$. This leaf node, $DTNU_{A \setminus A', U, t}$, is therefore characterized as $\{A', \emptyset, C', \emptyset\}$ and is a DTN. We add the constraints $a'_i \geq t, \forall a'_i \in A'$ and use a mixed integer linear programming solver \cite{cplex2009v12} to solve the DTN. If a solution is found, the execution time values for each $a_i' \in A'$ are stored and the leaf node's truth value is set to \textit{true.} Otherwise, it is set to \textit{false.} After a truth value is assigned to the leaf node, the truth propagation function defined in \S \ref{truthvalue} is called to logically infer truth value properties for parent nodes. 

Lastly, the search algorithm explores the tree in a depth-first manner. At each \dor, \wor and \and node, children nodes are visited in the order they are created. Once a child node is selected, its entire subtree will be processed by the algorithm before the other children are explored. Some simplifications made in the exploration are detailed in \S \ref{simplications} in the appendix.

\section{Tree Search Characteristics}
\label{tree-search-chars}

\subsection{Reactive scheduling during waits}
\label{instant-scheduling}
Some situations may arise when instant scheduling of a controllable timepoint is necessary as soon as an uncontrollable timepoint occurs to satisfy a constraint. We designate as a \textit{conjunct} a constraint relationship of the form $v_i - v_j \in [x,y]$ or $v_i \in [x,y]$, where $v_i, v_j$ are timepoints and $x, y, \in \rm I\!R$. We refer to a constraint where several conjuncts are linked by $\lor$ operators as a \textit{disjunct}. If at any given DTNU node in the tree there is an activated uncontrollable timepoint $u$ with the potential to occur during the next wait and there is at least one unscheduled controllable timepoint $a$ such that a conjunct of the form $u - a\in [0,y], y \geq 0$ 
is present in the constraints, a reactive wait strategy is considered that schedules $a$ as soon as $u$ occurs. Let $\Phi = \{\phi_1, \phi_2, ..., \phi_s\} \subset A$ be the complete set of unscheduled controllable timepoints for which there are conjunct clauses $u - \phi_i \in [0,y]$. 
We denote as $R_1, R_2, ..., R_m$ all possible combinations of elements taken from $\Phi$, including the empty set. As depicted in Figure \ref{fig:ts-structure-fig}, we account for potential reactive wait strategies by using a \wor node. The child node $\textit{AND}_{R_i}$ of the \wor node resulting from the combination $R_i$ has a reactive wait strategy for which all controllable timepoints in $R_i$ will be immediately executed at the moment $u$ occurs during the wait, if it does. If $u$ doesn't occur, no controllable timepoint is reactively scheduled during the wait.

\subsection{Wait action}
\label{wait}
When a wait decision of duration $\Delta_t$ is taken at time $t$ for a given DTNU node, two different categories of uncontrollable timepoints are considered to account for all transitional possibilities: \\

\begin{itemize}
    \item $ Z = \{\zeta_1, \zeta_2, ..., \zeta_l\}$ is a set of timepoints that could either happen during the wait, or afterwards, i.e. the end of the activation time interval for each $\zeta_i$ is greater than $t + \Delta_t$.
    \item $ H = \{\eta_1, \eta_2, ..., \eta_m\}$ is a set of timepoints that are certain to happen during the wait, i.e. the end of the activation time interval for each $\eta_i$ is less than or equal to $t + \Delta_t$.
\end{itemize}{}

There are $q = 2^l$ number of different possible combinations (empty set included) $V_1, V_2, ..., V_q$ for elements taken from $Z$. For each combination $V_i$, the set $\Lambda_i = H \cup V_i$ is created. The union $\bigcup\limits_{i=1}^{q} \Lambda_i$ refers to all possible combinations of uncontrollable timepoints which can occur by $t+ \Delta_t$. In figure \ref{fig:ts-structure-fig}, for each \and node, the combination $\Lambda_i$ leads to a DTNU sub-problem $DTNU_{O_i, P \cup \Lambda_i, t + \Delta_t}$ for which the uncontrollable timepoints in $\Lambda_i$ are considered to have occurred between $t$ and $t + \Delta_t$ in the schedule memory $S$. In addition, any potential controllable timepoint $\phi$ planned to be instantly scheduled in a reactive wait strategy $R_i$ in response to an uncontrollable timepoint $u$ in $\Lambda_i$ will also be considered to have been scheduled between $t$ and $t + \Delta_t$  in $S$. The only exception is when checking constraint satisfiability for the conjunct $u - \phi \in [0,y]$ which required the reactive scheduling, for which we assume $\phi$ executed at the same time as $u$, thus the conjunct is  considered satisfied automatically.

\subsection{Wait Eligibility and Period}
\label{waitperiod}

The way time is discretized is fundamental and holds direct implications on the search space explored and the capability of the algorithm to find TDC strategies. Longer waits make the search space smaller, but carry the risk of missing key moments where a decision is needed. On the other hand, smaller waits can make the search space too large to explore. We explain when the wait action is eligible, and how the wait duration is computed.

\paragraph{Eligibility}
At least one of the following criteria has to be met for a \wait node to be added as child of a \dor node: 
\begin{itemize}

    \item There is at least one activated uncontrollable timepoint for the parent DTNU node.
    
    \item There is at least one conjunct of the form $v \in [x,y]$, where $v$ is a timepoint, in the constraints of the parent DTNU node.
    
\end{itemize}
These criteria ensure that the search tree will not develop branches below \wait nodes when waiting is not relevant, \ie when a controllable timepoint necessarily needs to be scheduled. It also prevents the tree search from getting stuck in infinite \wait loop cycles.

\paragraph{Wait Period} We define the wait duration $\Delta_t$ at a given \dor node eligible for a wait dynamically by examining the updated constraint list $C'$ of the parent DTNU and the activation time intervals $B$ of its activated uncontrollable timepoints. Let $t$ be the current time for this DTNU node. The wait duration is defined by comparing $t$ to elements in $C'$ and $B$ to look for a minimum positive value defined by the 
following three rules:

\vspace{-0.5cm}

\subparagraph{First rule} For each activated time interval $u \in [x,y] $ in $B$, we select $x - t$ or $y - t$, whichever is smaller and positive, and we keep the smallest value $\delta_1$ found over all activated time intervals.

\vspace{-0.5cm}

\subparagraph{Second rule} For each conjunct $v \in [x,y] $ in $C'$, where $v$ is a timepoint, we select $x - t$ or $y - t$, whichever is smaller and positive, and we keep the smallest value $\delta_2$ found over all conjuncts.

\vspace{-0.5cm}

\subparagraph{Third rule} This rule is used to determine timepoints which need to be scheduled ahead of time by chaining constraints together. Intuitively, when a conjunct $v \in [x,y] $ is in $C'$, it means $v$ has to be executed when $t \in [x,y]$ to satisfy this conjunct.  However, $v$ could be linked to other timepoints by constraints which require them to happen before $v$. These timepoints could in turn be linked to yet other timepoints in the same way, and so on. The purpose of the third rule is to chain backwards to identify potential timepoints which start this chain and potential time intervals in which they need to be executed. The following mechanism is used: for each conjunct $v \in [x,y] $ in $C'$ found in 2), we apply a recursive backward chain function to both $(v, x)$ and $(v, y)$. We detail here how it is applied to $(v, x)$, the process being the same for $(v, y)$. Conjuncts of the form $v - v' \in [x', y'], x' \geq 0 $ in $C'$ are searched for. For each conjunct found, we add to a list two elements, $(v', x - x')$ and $(v', x - y')$. We also select $x - x' - t$ or $x - y' - t$, whichever is smaller and positive, as potential minimum candidate. The backward chain function is called recursively on each element of the list, proceeding the same way. We keep the smallest candidate $\delta_3$. Figure \ref{dynamicwait} in the appendix illustrates an application of this process. 

We set $\Delta_t = \min(\delta_1, \delta_2, \delta_3)$ as the wait duration. This duration is stored inside the \wait node.

\subsection{Truth Value Propagation}
\label{truthvalue}
In this section, we describe how truth attributes of nodes are related to each other. The truth attribute of a tree node represents its TDC controllability, and the relationships shared between nodes make it possible to define sound strategies. When a leaf node is assigned a truth attribute $\beta$, the tree search is momentarily stopped and a propagator function \texttt{PropagateTruth()} is called. This function recursively propagates $\beta$ onto upper parent nodes. A parent node $\boldsymbol{\omega}$ is selected recursively and we distinguish the following cases:

\begin{itemize}
    \item The parent $\boldsymbol{\omega}$ is a DTNU node or a \wait node: $\boldsymbol{\omega}$ is assigned $\beta$.
    \item The parent $\boldsymbol{\omega}$ is a \dor or \wor node: If $\beta = true$,  then $\boldsymbol{\omega}$ is assigned \true. If $\beta = false$ and all children nodes of $\boldsymbol{\omega}$ have \false attributes, $\boldsymbol{\omega}$ is assigned \false. Otherwise, the propagation stops.
    \item The parent $\boldsymbol{\omega}$ is an \and node: If $\beta = false$,  then $\boldsymbol{\omega}$ is assigned \false. If $\beta = true$ and all children nodes of $\boldsymbol{\omega}$ have \true attributes, $\boldsymbol{\omega}$ is assigned \true. Otherwise, the propagation stops.
\end{itemize}

After the propagation algorithm finishes, the tree search algorithm resumes where it was temporarily stopped. If a truth attribute has reached the root node of the tree, the tree search algorithm will be swiftly ended due to the branch cuts implemented in \S \ref{searchoptimizations} in the appendix. A \true attribute reaching the root node of the tree means a TDC strategy has been found. A \false attribute means none could be found. The pseudocode for the \texttt{PropagateTruth()} function is given in Algorithm \ref{truthvaluealgo} in the appendix.

\subsection{Constraint Propagation}
\label{tightbounds}

Decisions taken in the tree define when controllable timepoints are executed and also bear consequences on the execution time of uncontrollable timepoints. We explain here how these decisions are propagated into constraints, as well as the concept of `\textit{tight bound}'. Let $C'$ be the list of updated constraints for a DTNU node $\boldsymbol{\psi}$ for which the parent node is $\boldsymbol{\omega}$. We distinguish two cases. Either $\boldsymbol{\omega}$ is a \dor node and $\boldsymbol{\psi}$ results from the execution of a controllable timepoint $a_i$, or $\boldsymbol{\omega}$ is an \and node and $\boldsymbol{\psi}$ results from a wait of $\Delta_t$ time units. In the first case, let $t$ be the execution time of $a_i$. The updated list $C'$ is built from the constraints of the parent DTNU of $\boldsymbol{\psi}$ in the tree. If a conjunct contains $a_i$ and is of the form $a_i \in [x,y]$, this conjunct is replaced with \textit{true} if $t \in [x,y]$, \textit{false} otherwise. If the conjunct is of the form $v_j - a_i \in [x,y]$, we replace the conjunct with $v_j \in [t + x, t + y]$. The other possibility is that $\boldsymbol{\psi}$ results from a wait of $\Delta_t$ time at time $t$, with a reactive wait strategy $R_j$. In this case, the new time is $t + \Delta_t$ for $\boldsymbol{\psi}$. As a result of the wait, some uncontrollable timepoints $u_i \in \Lambda _i$ may occur, and some controllable timepoints $a_i \in R_j$ may be executed reactively during the wait. Let $v_i \in \Lambda_i \cup R_j$ be these timepoints occurring during the wait. The execution time of these timepoints is considered to be in [$t$, $t+\Delta_t$]. For uncontrollable timepoints $u_i' \in \Lambda_i' \subset \Lambda_i$ for which the activation time ends at $t+\Delta_{t_i}'< t+\Delta_t$, and potential controllable timepoints $a_i'$ instantly reacting to these uncontrollable timepoints, the execution time is further reduced and considered to be in $[t, t+\Delta_{t_i}']$. We define a concept of \textit{tight bound} to update constraints which restricts time intervals in order to account for all possible values $v_i$ can take between $t$ and $t+\Delta_t$. For all conjuncts $v_j - v_i \in [x,y]$, we replace the conjunct with $v_j \in [t + \Delta_t + x, t + y]$. Intuitively, this means that since $v_i$ can happen at the latest at $t + \Delta_t$, $v_j$ can not be allowed to happen before $t + \Delta_t + x$. Likewise, since $v_i$ can happen at the earliest at $t$,  $v_j$ can not be allowed to happen after $t + y$. Finally, if $t+ \Delta_t + x > t + y$, the conjunct is replaced with \false. Also, the process can be applied recursively in the event that $v_j$ is also a timepoint that occurred during the wait, in which case the conjunct would be replaced by \true or \textit{false}. In any case, any conjunct obtained of the form $a_j \in [x',y']$ is replaced with \false if $t + \Delta_t > y'$. Finally, if all conjuncts inside a disjunct are set to \false by this process, the constraint is violated and the DTNU is no longer satisfiable.
\section{Learning-based Heuristic}
\label{network}

We present our learning model and explain how it provides tree search heuristic guidance. Our learning architecture originates from \cite{gilmer2017neural}. It uses message passing rules allowing neural networks to process graph-structured inputs where both vertices and edges possess features. Authors of this architecture carry out node classification experiments in quantum chemistry and achieve state-of-the-art results on a molecular property prediction benchmark. Here, we first define a way of converting DTNUs into graph data. Then, we process the graph data with our MPNN and explain how the output is used to guide the tree search. 

Let $\Gamma = \{A,U,C,L\}$ be a DTNU. We start by explaining how we turn $\Gamma$ into a graph $\mathcal{G} = (\mathcal{V}, \mathcal{E})$. First, we convert all time values from absolute to relative with the assumption the current time for $\Gamma$ is $t=0$. We search all converted time intervals $[x_i,y_i]$ in $C$ and $L$ for the highest interval bound value $d_{max}$, \ie the farthest point in time. We then proceed to normalize every time value in $C$ and $L$ by dividing them by $d_{max}$. As a result, every time value becomes a real number between $0$ and $1$. Next, we convert each controllable timepoint $a \in A$ and uncontrollable timepoint $u \in U$ into graph nodes with corresponding \textit{controllable} or \textit{uncontrollable} node features. The time constraints in $C$ and contingency links in $L$ are expressed as edges between nodes with $10$ different edge distance classes ($0:[0,0.1)$, $1:[0.1,0.2)$, ...,  $9:[0.9,1]$). We also use additional edge features to account for edge types (constraint, disjunction, contingency link, direction sign for lower and upper bounds). Moreover, intermediary nodes are used with a distinct node feature in order to map possible disjunctions in constraints and contingency links. We also add a \wait node with a distinct node feature which implicitly designates the act of waiting a period of time. The graph conversion of DTNU $\gamma$ is characterized by three elements: the matrix of all node features $X_{v}$, the adjacency matrix of the graph $X_{e}$ and the matrix of all edge features $X_{w}$.

Let $f$ be the mathematical function for our MPNN and $\theta$ its set of parameters. Our function $f$ stacks $5$ graph convolutional layers from \cite{gilmer2017neural} coupled with the $\mathrm{ReLU}(\cdot) = \max(0,\cdot)$ piece-wise activation function \cite{glorot2011deep}. The $\texttt{sigmoid}$ function 
$\sigma(\cdot) = \frac{1}{1+\exp(-\cdot)}$ 
is then used to obtain a list of probabilities $\pi$ over all nodes in $\mathcal{G}$ : $f_{\theta}(X_{v}, X_{e}, X_{w}) = \pi$. 
The probability of each node $v$ in $\pi$ corresponds to the likelihood of transitioning into a TDC DTNU from the original DTNU $\Gamma$ by taking the action corresponding to $v$. If $v$ represents a controllable timepoint $a$ in $\Gamma$, its corresponding probability in $\pi$ is the likelihood of the sub-DTNU resulting from the execution of $a$ being TDC. If $v$ represents a \wait decision, its probability refers to the likelihood of the \wait node having a \true attribute, \ie the likelihood of all children DTNUs resulting from the wait being TDC (with the wait duration rules set in \S \ref{waitperiod}). We call these two types of nodes \textit{active} nodes. Otherwise, if $v$ is another type of node, its probability is not relevant to the problem and ignored. Our MPNN is trained on DTNUs generated and solved in \S \ref{randomized-tree-search} only on active nodes by minimizing the cross-entropy loss: $$\frac{1}{m} \sum\limits_{i=1}^{m} \sum\limits_{j=1}^{q} - Y_{ij} \log (f_{\theta}(X_i)_j) - (1 - Y_{ij}) \log (1-f_{\theta}(X_i)_j)$$ Here $X_i = (X_{i_v}, X_{i_e}, X_{i_w})$ is DTNU number $i$ among a training set of $m$ examples, $Y_{ij}$ is the TDC controllability (1 or 0) of active node $j$ for DTNU number $i$.

Lastly, the MPNN heuristic is used in the following way in the tree search. Once a \dor node is reached, the parent DTNU node is converted into a graph and the MPNN $f$ is called upon the corresponding graph elements $X_{v}, X_{e}, X_{w}$. Active nodes in output probabilities $\pi$ are then ordered by highest values first, and the tree search visits the corresponding children tree nodes in the suggested order, preferring children with higher likelihood of being TDC first. 

\section{Randomized Simulations for Heuristic Training}
\label{randomized-tree-search}

We leverage a learning-based heuristic to guide the tree search and increase its effectiveness. A key component in learning-based methods is the annotated training data. We generate such data in automatic manner by using a DTNU generator to create random DTNU problems and solving them with a modified version of the tree search. We store results and use them for training the MPNN. We detail now our data generation strategy.

For training purposes, we create DTNUs which have a number of controllable timepoints ranging from 10 to 20 and a number of uncontrollable timepoints ranging from 1 to 3. The DTNUs are generated in the following way. For interval bounds of constraint conjuncts or contingency links, we always  randomly generate real numbers within $[0,100]$. We restrict the number of conjuncts inside a disjunct to 5 at most. A random number $n_1 \in [10,20]$  of controllable timepoints and a random number $n_2 \in [1,3]$ of uncontrollable timepoints are selected. Each uncontrollable timepoint is randomly linked to a different controllable timepoint with a contingency link. Next, we iterate over the list of timepoints, and for each timepoint $v_i$ not appearing in constraints or contingency links, we add a disjunct for which at least one conjunct constrains $v_i$. The type of conjunct is selected randomly from either a \textit{distance} conjunct $v_i - v_j \in [x,y]$ or a \textit{bounded} conjunct $v_i\in [x,y]$. On the other hand, if $v_i$ was already present in the constraints or contingency links, we add a disjunct constraining $v_i$ with only a $20\%$ probability.
In order to solve these random DTNUs, we modify the tree search as follows. 
For a DTNU $\Gamma$, the first \dor child node is developed as well as its children $\psi_1, \psi_2, ..., \psi_n \in \Psi$. The modified tree search will explore each $\psi_i$ multiple times ($\nu$ times at most), each time with a timeout of $\tau$ seconds. 
Here we set $\nu = 25$ and $\tau = 3$. For each exploration of $\psi_i$, children nodes of any \dor node encountered in the corresponding subtree are explored randomly each time. If $\psi_i$ is proved to be either TDC or non-TDC during an exploration, the next explorations of the same child $\psi_i$ are called off and the truth attribute $\beta_i$ of $\psi_i$ is updated accordingly. The active node number $k$, corresponding to the decision leading to $\psi_i$ from DTNU $\Gamma$'s \dor node, is updated with the same value, \ie $Y_{k} = \beta_i$. However, if every exploration times out, $\psi_i$ is assumed non-TDC and $Y_k$ is set to \textit{false}. Once each $\psi_i$ has been explored, the pair $\langle \mathcal{G}(\Gamma), (Y_1, Y_2, ..., Y_n)\rangle$ is stored in the training set, where $\mathcal{G}(\Gamma)$ is the application of the graph conversion of $\Gamma$ 
described in \S\ref{network}.

The assumption of non-TDC controllability for children nodes for which all explorations time out is good enough. The output of the MPNN is a probability for each child node of the \dor node of the input DTNU. These children nodes are visited in the suggested order  when the heuristic is active, until one is found to be TDC, and no child is ever discarded. The trained MPNN will tend to give higher probabilities for children nodes for which explorations often found a TDC strategy before timeout, and lower probabilities for ones where explorations often ended up with a timeout.

\section{Strategy Execution}
\label{strategy-exec}

A strategy found by the tree search for a DTNU $\Gamma$ is sound and guarantees constraint satisfiability if executed 
in the following manner.
Let $\mathcal{Q}$ be the system interacting with the environment,  executing controllable timepoints and observing how uncontrollable timepoints unfold. At each DTNU node in the tree, $\mathcal{Q}$ will move on to the child \dor node. The child node $\psi_i$ of the \dor node which was found by the strategy to have a \true attribute is selected. If $\psi_i$ is a DTNU node, $\mathcal{Q}$ executes the corresponding controllable timepoint $a_i$ and moves on to $\psi_i$. On the other hand, if $\psi_i$ is a \wait node, $\mathcal{Q}$ moves on to $\psi_i$, reads the wait duration time $\Delta_t$ stored in $\psi_i$ and moves on to the child \wor node. The child node \textit{AND}$_{R_j}$ of the \wor node which has a \true attribute is selected, and $\mathcal{Q}$ will wait $\Delta_t$ time units with the reactive wait strategy $R_j$. After the wait is over, $\mathcal{Q}$ observes the list of all uncontrollable timepoints $\Lambda_i$ which occurred, deduces which DTNU child node of the \textit{AND}$_{R_j}$ node it transitioned into, and moves on to that node.

By following these guidelines, the final tree node $\mathcal{Q}$ 
transitions into is necessarily a leaf node with a \true attribute, \ie a node for which all constraints are satisfied. This is due to the fact that for \dor and \wor nodes $\mathcal{Q}$ visits, $\mathcal{Q}$ chooses to transition into a child node with a \true attribute. For \and nodes $\mathcal{Q}$ visits, all children DTNU nodes have a \true attribute, so $\mathcal{Q}$ transitions into a child node with a \true attribute regardless of how uncontrollable timepoints unfold.

\section{Related Works}
The use of learning-based heuristics has recently become increasingly popular  for planning, combinatorial and network modeling problems. Recent works applied to network modeling and routing problems include \cite{rusek2019unveiling},  \cite{chen2018deep}, \cite{xu2018experience}, \cite{Kool2018}. Recently, GNNs have become a popular extension of CNNs. Essentially, their ability to represent problems with a graph structure and the resulting node permutation invariance makes them convenient for some applications. We refer the reader to \cite{wu2019comprehensive} for a complete survey on GNNs. In combinatorial optimization, GNNs can benefit both approximate and exact solvers. In \cite{ZhuwenLi2018}, authors combine tree search, GNNs and a local search algorithm to achieve state-of-the-art results for approximate solving of NP-hard problems such as the maximum independent set problem. On the other hand, \cite{gasse2019exact} use a GNN for branch and bound variable selection for exact solving of NP-hard problems and achieve superior results to previous learning approaches. In path-planning problems with NP-hard constraints, \cite{osanlou2019optimal} use a GNN to predict an upper bound for a branch and bound solver and outperform an A*-based planner coupled with a problem-suited handcrafted heuristic. Lastly, \cite{Ma2018} call a GNN for the selection of a planner inside a portfolio for STRIPS planning problems and outperform the leading learning-based approach which was based on a CNN \cite{sievers2019deep}. In most works, GNNs seem to offer generalization to bigger problems than they are trained on. Results from our experiments are in line with this observation.

\section{Experiments}

\begin{figure}[tb]
\renewcommand{\captionfont}{\small}
\centering
\includegraphics[scale=0.56]{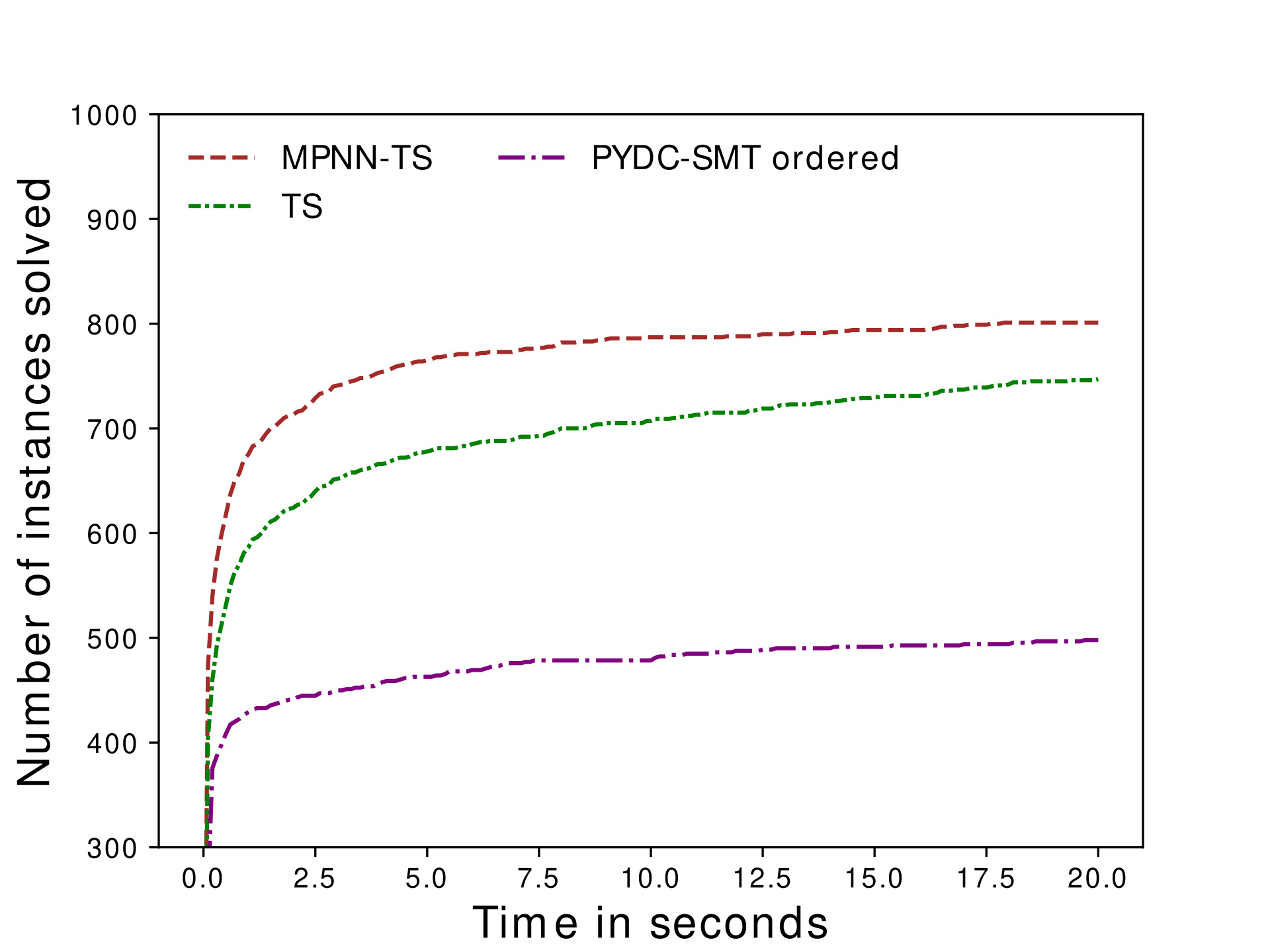}
\caption{\textbf{Experiments on \cite{cimatti2016dynamic}'s benchmark from which DTNs and STNs have been removed.} The X-axis represents the allocated time in seconds and the Y-axis the number of instances in the benchmark each solver can solve within the corresponding allocated time. Timeout is set to 20 seconds per instance.}
\label{aless-bench}
\vspace{-0.4cm}
\end{figure}

\begin{figure}[tb]
\renewcommand{\captionfont}{\small}
\centering
\includegraphics[scale=0.56]{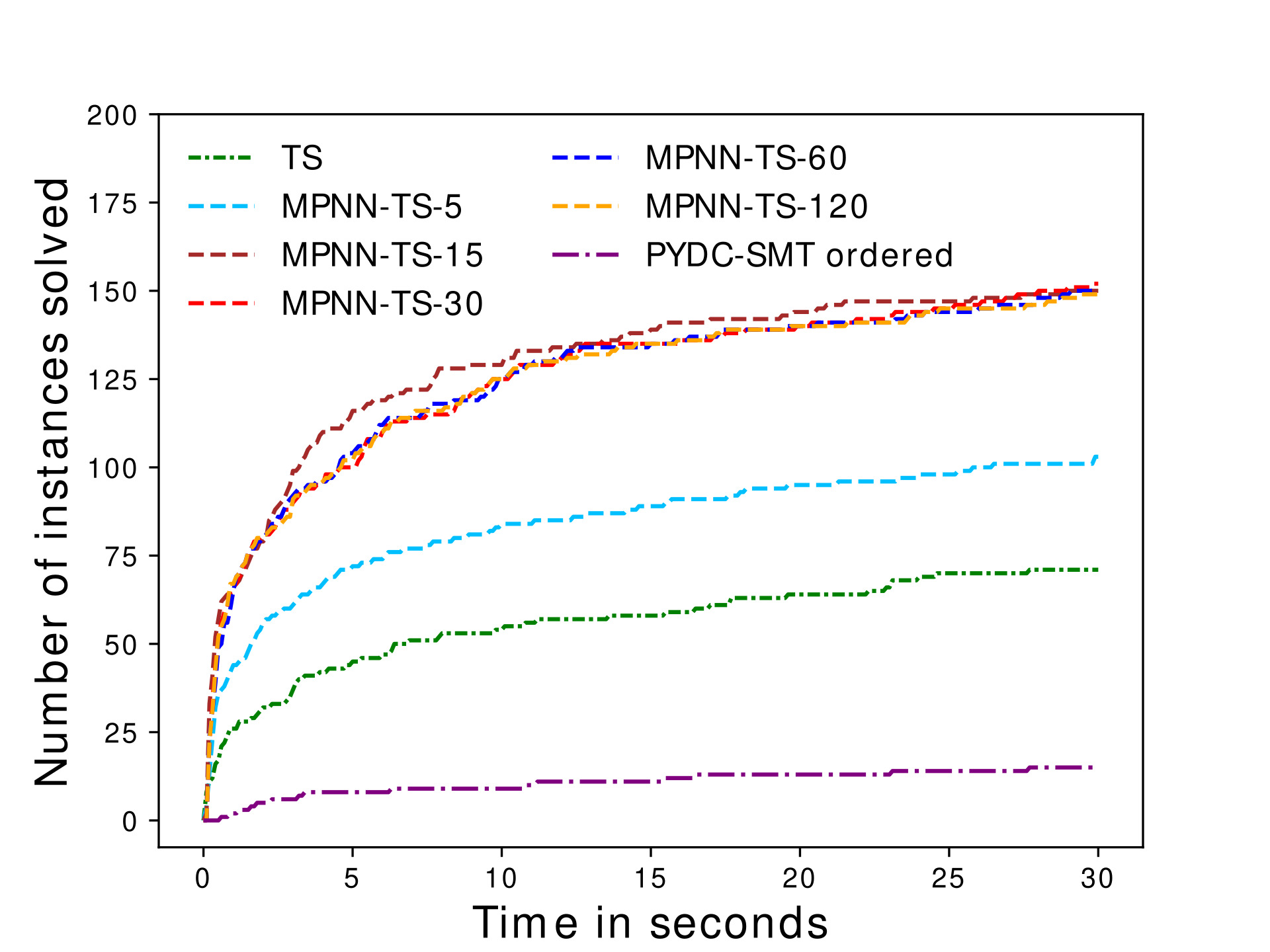}
\caption{\textbf{Experiments on benchmark $\boldsymbol{B_1}$.} Axes are the same as in figure \ref{aless-bench}. Timeout is set to 30 seconds per instance.}
\label{10-20}
\vspace{-0.5cm}
\end{figure}

We carry out experiments to evaluate the efficiency of the proposed tree search approach and the effect of the MPNN's guidance. We also compare these methods to a DC solver from \cite{cimatti2016dynamic}. TDC is a subset of DC and a more restrictive form of controllability: non-TDC controllability does not imply non-DC controllability. In that sense, a TDC solver can be expected to offer better performance that a DC counterpart in exchange for potentially being unable to find a strategy when a DC algorithm would. In this section, we refer to the tree search algorithm as TS, the tree search algorithm guided by the trained MPNN up to the $15^{th}$ (respectively $X^{th}$) \dor node depth-wise in the tree as MPNN-TS (respectively MPNN-TS-X) and the most efficient DC solver from \cite{cimatti2016dynamic} as PYDC-SMT ordered.

First, we use the benchmark in the experiments of  \cite{cimatti2016dynamic} from which we remove DTNs and STNs. %
We compare TS, MPNN-TS and PYDC-SMT on the resulting benchmark which is comprised of $290$ DTNUs and $1042$ STNUs. Here, Limiting the maximum depth use of the MPNN to $15$ offers a good trade off between guidance gain and cost of calling the heuristic. Results are given in Figure~\ref{aless-bench}. We observe that TS solves roughly $50\%$ more problem instances than PYDC-SMT within the allocated time ($20$ seconds). In addition, TS solves $56\%$ of all instances while the remaining ones time out. Among solved instances, a strategy is found for $89\%$ and the remaining $11\%$ are proved non-TDC. On the other hand, PYDC-SMT solves $37\%$ of all instances. A strategy is found for $85\%$ of PYDC-SMT's solved instances while the remaining $15\%$ are proved non-DC. Finally, out of all instances PYDC-SMT solves, TS solves $97\%$ accurately with the same conclusion, \ie TDC when DC and non-TDC when non-DC. The use of the heuristic leads to an additional $+6\%$ problems solved within the allocated time. We argue this small increase is essentially due to the fact that most problems solved in the benchmark are small-sized problems with few timepoints which are solved quickly. 

For further evaluation of the heuristic, we create new benchmarks using the DTNU generator described in \S \ref{randomized-tree-search} with varying number of timepoints. These benchmarks contain fewer DTNU instances which are quick to solve and more harder instances. Each benchmark contains 500 randomly generated DTNUs which have $1$ to $3$ uncontrollable timepoints. Moreover, each DTNU has 10 to 20 controllable timepoints in the first benchmark $B_1$, 20 to 25 in the second benchmark $B_2$ and 25 to 30 in the last benchmark $B_3$. Experiments on $B_1$, $B_2$ and $B_3$ are respectively shown in figure \ref{10-20}, \ref{20-25-b} (in the appendix) and \ref{25-30}. We note that for all three benchmarks, no solver ever proves non-TDC or non-DC controllability before timing out due to the larger size of these problems.

\begin{figure}[tb]
\renewcommand{\captionfont}{\small}
\centering
\includegraphics[scale=0.56]{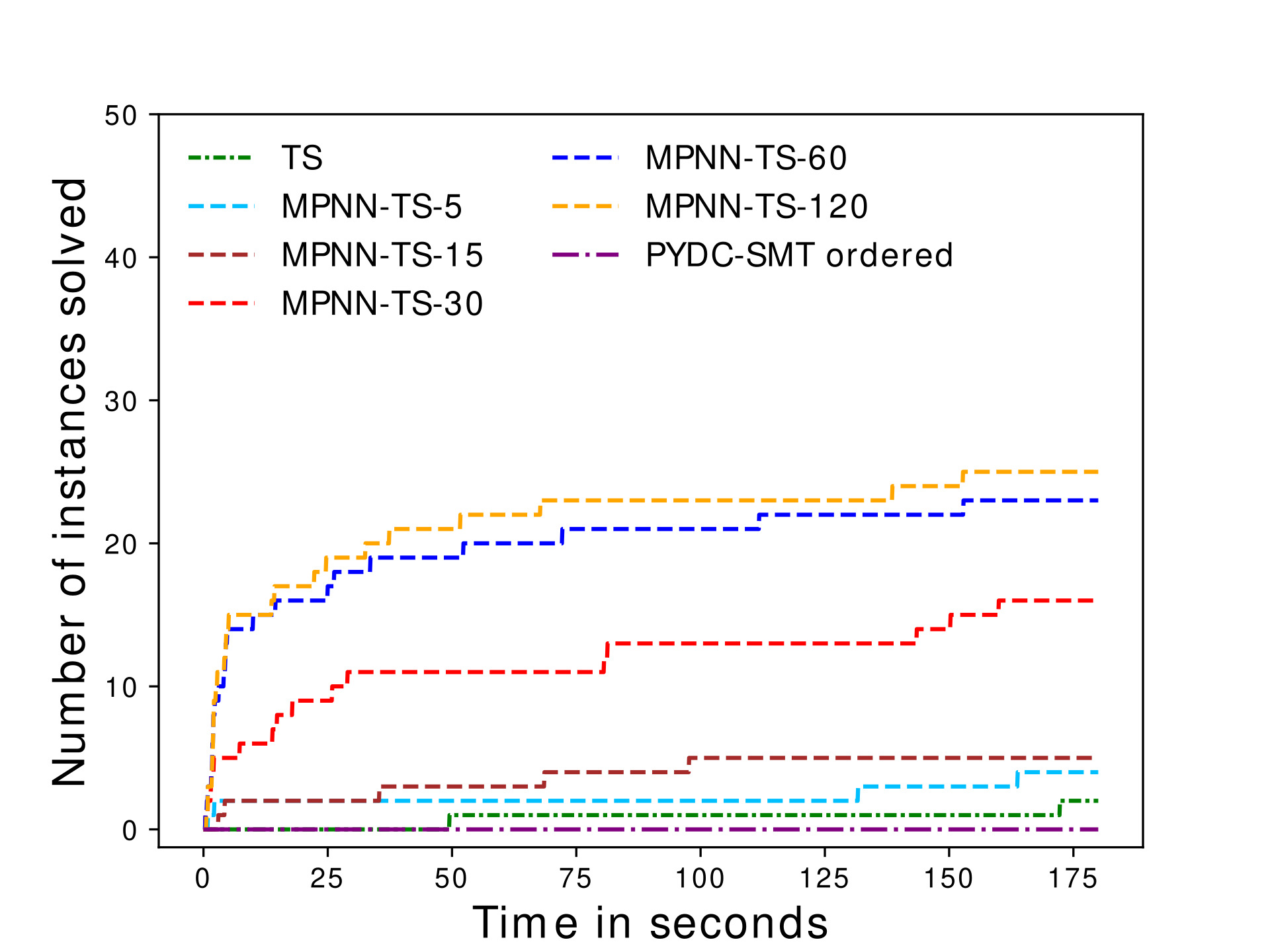}
  \caption{\textbf{Experiments on benchmark $\boldsymbol{B_3}$.} Axes are the same as in figure \ref{aless-bench}. Timeout is set to 180 seconds.}
\label{25-30}
\vspace{-0.5cm}
\end{figure}

In these benchmarks, PYDC-SMT does not perform well on $B_1$ and cannot solve any instance on $B_2$ and $B_3$. TS does not perform well on $B_2$ and only solves 2 instances on $B_3$. However, we see a significantly higher gain from the use of the MPNN for TS, varying with the maximum depth use. At best depth use, the gain is $+91\%$ instances solved for $B_1$, $+980\%$ instances solved for $B_2$ and $+1150\%$ instances solved for $B_3$. The more timepoints instances have, the more worthwhile heuristic guidance appears to be. Indeed, the optimal maximum depth use of the MPNN in the tree increases with the problem size: $15$ for $B_1$, $60$ for $B_2$ and $120$ for $B_3$. We argue this is due to the fact that more timepoints results in a wider search tree overall, including in deeper sections where heuristic use was not necessarily worth its cost for smaller problems. Furthermore, the MPNN is trained on randomly generated DTNUs which have 10 to 20 controllable timepoints. The promising gains shown by experiments on $B_2$ and $B_3$ suggest generalization of the MPNN to bigger problems than it is trained on.

The tree search approach presented in this work presents a good trade off between search completeness and effectiveness: almost all examples solved by PYDC-SMT from \cite{cimatti2016dynamic}'s benchmark are solved with the same conclusion, and many more which could not be solved are. Moreover, the TDC approach scales up better to problems with more timepoints, and the tree structure allows the use of learning-based heuristics. Although these heuristics are not key to solving problems of big scales, our experiments suggest they can still provide a high increase in efficiency.

\section{Conclusion}
We introduced a new type of controllability, time-based dynamic controllability (TDC), and a tree search approach for solving disjunctive temporal networks with uncertainty (DTNU) in TDC. Strategies are built by discretizing time and exploring different decisions which can be taken at different key points, as well as anticipating how uncontrollable timepoints can unfold. We defined constraint propagation rules which ensure soundness of strategies found. We showed that the tree search approach is able to solve DTNUs in TDC more efficiently than the state-of-the-art dynamic controllability (DC) solver, PYDC-SMT, with almost always the same conclusion. Lastly, we created MPNN-TS, a solver which combines the tree search with a heuristic function based on message passing neural networks (MPNN) for guidance. The MPNN is trained with a self-supervised strategy based on a variant of the tree search algorithm. The use of the MPNN allows significant improvement of the tree search on harder DTNU problems, notably on DTNUs of bigger size than those used for training the MPNN.

\bibliographystyle{aaai}
\bibliography{prl}

\clearpage

\onecolumn

\section{Appendix}

\subsection{Plots}

\bigskip

\begin{figure*}[!htb]
    \renewcommand{\captionfont}{\small}
	\centering
	\begin{subfigure}[t]{0.47\textwidth}
		\centering
		\includegraphics[width=0.999\linewidth]{figures/aless300.jpg}
		\caption{Experiments on \cite{cimatti2016dynamic}'s benchmark from which the DTNs and STNs have been removed. The X-axis represents the allocated time in seconds and the Y-axis the total number of instances that each solver can solve within the corresponding allocated time. Timeout is set to 20 seconds per instance.}\label{alless-b}		
	\end{subfigure}
	\quad
	\begin{subfigure}[t]{0.47\textwidth}
		\centering
		\includegraphics[width=0.999\linewidth]{figures/10-20.jpg}
		\caption{Experiments on benchmark $B_1$. Axes are the same as in figure \ref{alless-b}. Timeout is set to 30 seconds per instance.}\label{10-20-b}
	\end{subfigure}
\newline
	\centering
	\begin{subfigure}[t]{0.47\textwidth}
		\centering
		\includegraphics[width=0.999\linewidth]{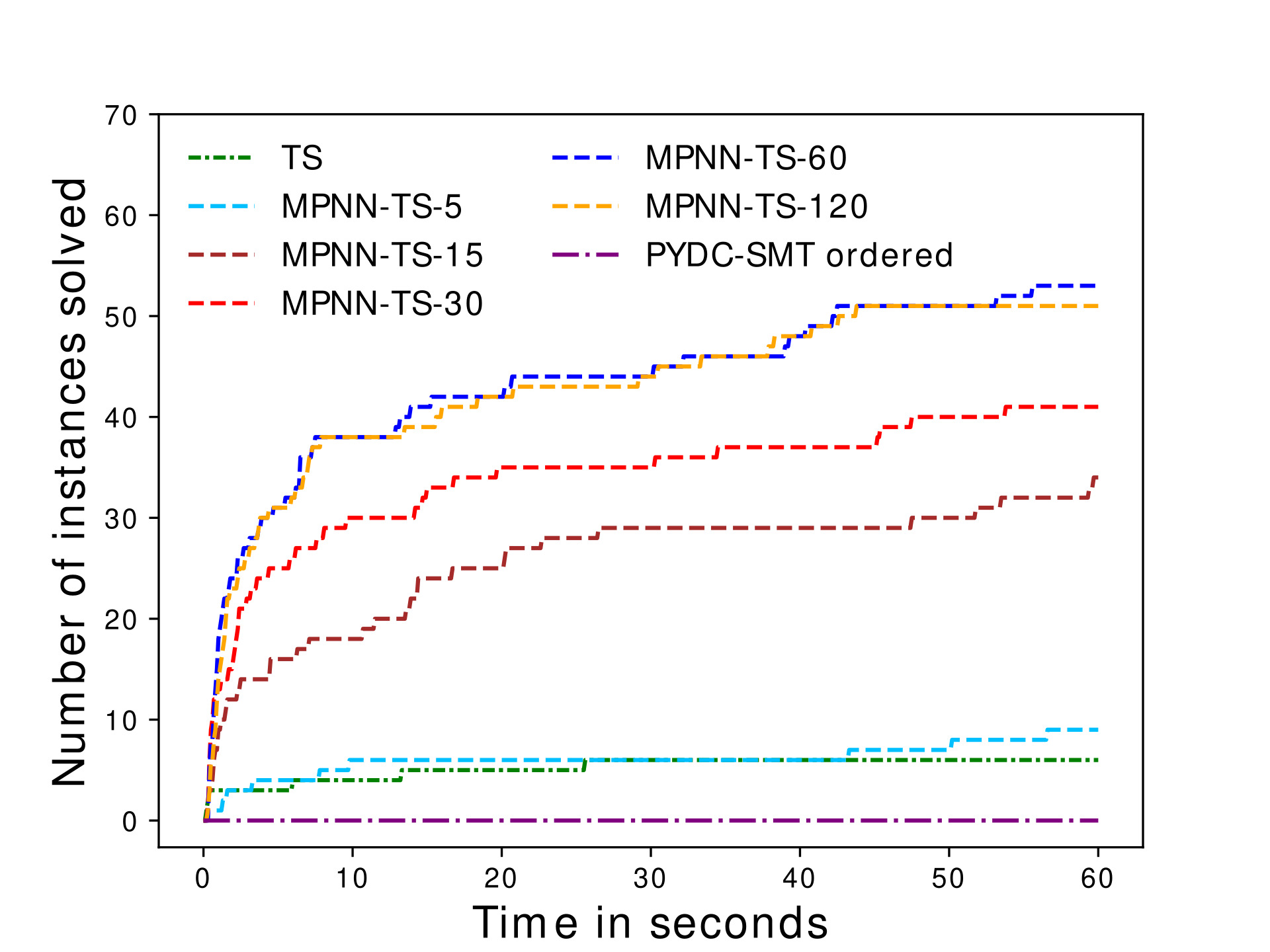}
		\caption{Experiments on benchmark $B_2$. Axes are the same as in figure \ref{alless-b}. Timeout is set to 60 seconds per instance.}
		\label{20-25-b}
	\end{subfigure}
	\quad
	\begin{subfigure}[t]{0.47\textwidth}
		\centering
		\includegraphics[width=0.999\linewidth]{figures/25-30.jpg}
		\caption{Experiments on benchmark $B_3$. Axes are the same as in figure \ref{alless-b}. Timeout is set to 180 seconds per instance.}
		\label{25-30-b}
	\end{subfigure}
	\caption{\textbf{Summary of experiments on benchmarks}}\label{fig-summary}
\end{figure*}

\clearpage

\twocolumn

\bigskip

\subsection{Simplified Example}

\begin{figure}[H]
\renewcommand{\captionfont}{\small}
\centering
\includegraphics[scale=0.75]{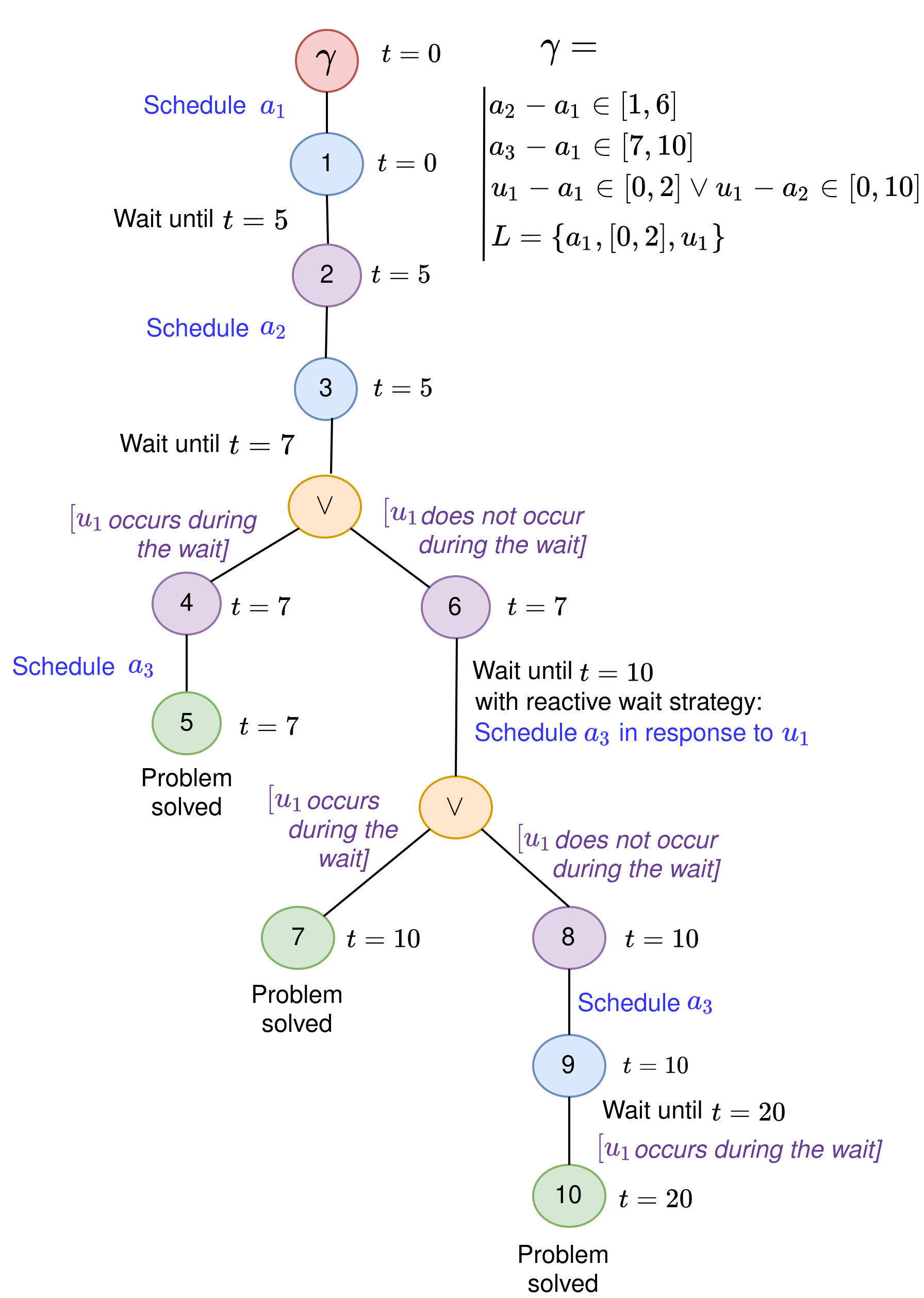}
\caption{\textbf{Simplified TDC strategy of a DTNU $\Gamma$.} For space reasons, we only give a summarized copy of the strategy found. Branches leading to unsolved cases are excluded, and we do not include \dor, \wor, \and and \wait nodes. The node $\gamma$ is the original DTNU. Other nodes are sub-DTNUs, except the $\lor$ node which aims to list transitional possibilities, and should be interpreted in the figure as an \and node.}
\label{alessandroexample}
\end{figure}

Figure \ref{alessandroexample} is a simplified example of a TDC strategy of the example DTNU from \cite{cimatti2016dynamic}.

\vfill\null

\subsection{Truth Value Propagation Algorithm}

We present in this section Algorithm \ref{truthvaluealgo}. This algorithm is called to propagate a truth value in the tree. The propagation is done in an ascending way: truth values are inferred from the leaves of the tree towards the root.

\begin{algorithm}[H]
\small
\caption{Truth Value Propagation}\label{truthvaluealgo}
\begin{algorithmic}[1]
\Function{propagateTruth(TreeNode $\psi$)}{}
    \State $\omega \gets $ parent($\psi$) \Comment{$^{1}*$}
    \If {$\omega = null$}
        \State \textbf{return}
    \EndIf
    \If {isDTNU($\omega$) or isWAIT($\omega)$}  \Comment{$^{2}*$}
	 	\State $\omega .truth \gets \psi .truth$
	 	\State propagateTruth($\omega$)
	\ElsIf {isOR($\omega$)}   \Comment{$^{3}*$}
	    \If {$\psi .truth = True$}
	        \State $\omega .truth \gets True$
	 	    \State propagateTruth($\omega$)
	 	\Else 
	 	    \If{$\forall \sigma_i, \sigma_i.truth = False$} \Comment{$^{4}*$}
	 	        \State $\omega .truth \gets False$
	 	        \State propagateTruth($\omega$)
	 	    \EndIf
	 	\EndIf
	
	\ElsIf {isAND($\omega$)} \Comment{$^{5}*$}
	    \If {$\psi .truth = False$}
	        \State $\omega .truth \gets False$
	 	    \State propagateTruth($\omega$)
	 	\Else 
	 	    \If{$\forall \sigma_i, \sigma_i.truth = True$} \Comment{$^{4}*$}
	 	        \State $\omega .truth \gets True$
	 	        \State propagateTruth($\omega$)
	 	    \EndIf
	 	\EndIf
	\EndIf

\EndFunction

\end{algorithmic}
\footnotemark{$*$ parent($x$): Returns the parent node of $x$, $null$ if none.\\}
\footnotemark{$*$ isDTNU($x$): Returns \True if $x$ is a DTNU node, \False otherwise; isWait($x$): Returns \True if $x$ is a \textit{WAIT} node, \False otherwise.\\}
\footnotemark{$*$ isOR($x$): Returns \True if $x$ is an \dor or \wor node, \False otherwise.\\}
\footnotemark{$*$ $\sigma_i$: Child number $i$ of $\omega$. For a \dor or \wor node, in the case where $\psi$ is \false but not all other children of $\omega$ are \false the propagation stops. Likewise, for an \and node and in the case where $\psi$ is \true but not all other children of $\omega$ are \true, the propagation stops.\\}
\footnotemark{$*$ isAnd($x$): Returns \True if $x$ is an \textit{AND} node, \False otherwise.\\}
\end{algorithm}

\clearpage

\subsection{Tree Search Algorithm}
We give the simplified pseudocode for the tree search in Algorithm \ref{tsalgo}
\begin{algorithm}[H]
\small
\caption{Tree Search}\label{tsalgo}
\begin{algorithmic}[1]

\Function{explore($\psi$)}{}
    
    \If {$parent(\psi).truth \neq unknown$}
        \State \textbf{return}
    \EndIf

    \If {isDTNU($\psi$)}
    
        \State updateConstraints($\psi$) \Comment{$^{6}*$}
    
        \If{IsLeaf($\psi$)} \Comment{$^{7}*$}
            \State propagateTruth($\psi$)
            \State \textbf{return}
        \EndIf
        
        \State Create \dor child $\psi'$
        \State explore($\psi'$)

    \EndIf
    
    \If {isOR($\psi$)}
        \State Create list of all children $\Psi'$ \Comment{$^{8}*$}
        \For{$\psi'\in \Psi' $}
            \State explore($\psi'$)
        \EndFor
    \EndIf

    \If {isAND($\psi$)}
        \State Create list of all children $\Psi'$ \Comment{$^{9}*$}
        \For{$\psi'\in \Psi' $}
            \State explore($\psi'$)
        \EndFor
    \EndIf
    
    \If {isWAIT($\psi$)}
        \State create \wor child $\psi'$
        \State explore ($\psi'$)
    \EndIf
    
\EndFunction

\Function{main(DTNU $\gamma$)}{}
    \State explore($\gamma$)
    \If {$\gamma . truth = True$}
        \State \textbf{return} $True$
    \Else
        \State \textbf{return} $False$
    \EndIf
\EndFunction

\end{algorithmic}
\footnotemark{$*$  updateConstraints($x$): Updates the constraints of DTNU node $x$.}

\footnotemark{$*$ isLeaf($x$): Sets the truth value of $x$ to \true and returns \true if all constraints are satisfied. Sets the truth value to \false and returns \true if a constraint is violated. If no truth value can be inferred at this stage with the updated constraints, a second check is run to determine if all uncontrollable timepoints have occurred. In this case, the corresponding DTN is solved, the truth value of $x$ is updated accordingly, and the function returns \true. Otherwise, if no logical outcome can be inferred for the current state of the constraints and there remains at least one uncontrollable timepoint, this function returns \false.}

\footnotemark{$*$ If this is a \dor node, the list $\Psi'$ contains all the children DTNU nodes resulting from either the decision of scheduling a controllable timepoint, or the \wait node resulting from a wait if available. If this is a \wor node,  $\Psi'$ contains all \textit{AND}$_{R_j}$ nodes, each of which possess a reactive wait strategy $R_j$ }

\footnotemark{$*$ Here, the list $\Psi'$ contains all DTNUs resulting from all possible combinations $\Lambda_1, \Lambda_2, ..., \Lambda_q$ of uncontrollable timepoints which have the potential to occur during the current wait.}

\end{algorithm}

\vfill\null

\subsection{Wait Period}

The following figure gives an example of the third rule used to compute a wait duration.

\begin{figure}[tb]
\renewcommand{\captionfont}{\small}
\centering
\includegraphics[scale=0.79]{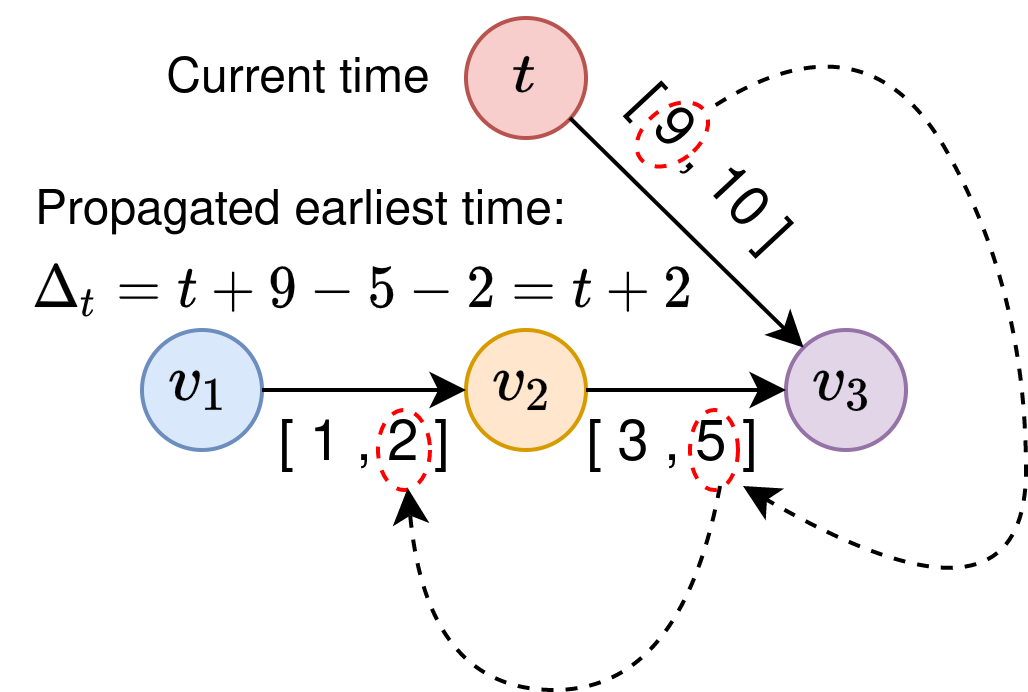}
\caption{\textbf{Application of the $3^{rd}$ rule to determine a wait duration.} Variables $v_1$, $v_2$ and $v_3$ are timepoints. Here, $v_2$ is constrained to execute in the time interval $[1,2]$ after $v_1$, $v_3$ in $[3,5]$ after $v_2$ and $v_3$ in $[t+9, t+10]$. It is suggested not to wait longer than $2$ units of time at $t$: an execution of $v_1$ at $t+2$, followed by an execution of $v_2$ at $t+4$ opens a window of opportunity for $v_3$ to execute at $t+9$.}
\label{dynamicwait}
\end{figure}

\subsection{Optimization Rules}
\label{searchoptimizations}
\label{simplications}
The following rules are added to make branch cuts when possible.

\paragraph{\textit{Constraint Check.}} When a DTNU node is explored and the updated list of constraints $C'$ is built according to \S \ref{tightbounds}, if a disjunct is found to be \false, $C'$ will no longer be satisfiable. All the subtree which can be developed from the DTNU will only have leaf nodes for which this is the case as well. Therefore, the search algorithm will not develop this subtree.
    
\paragraph{\textit{Symmetrical subtrees.}} Some situations can lead to the development of the exact same subtrees. A trivial example, for a given DTNU node at a time $t$, is the order in which a given combination of controllable timepoints $a_1, a_2, ..., a_k$ is taken before taking a wait decision. Regardless of what order these timepoints are explored in the tree before moving to a \wait node, they will be considered executed at time $t$. Therefore, when taking a wait decision, it is checked that all preceding controllable timepoints executed before the previous wait are a combination of timepoints that has not been not tested yet.
    
\paragraph{\textit{Truth Checks.}}  Before exploring a new node for which the truth attribute is set to \textit{unknown}, the truth attribute of the parent node is also checked. The node is only developed if the parent node's truth attribute is set to \textit{unknown}. In this manner, when children of a tree node are being explored depth-first and the exploration of a child node leads to the assignment of a truth value to the tree node, the remaining unexplored children can be left unexplored.

\subsection{Implementation details}
\label{implementation-details}
\begin{figure}[tb]
\renewcommand{\captionfont}{\small}
\centering
\includegraphics[scale=0.7]{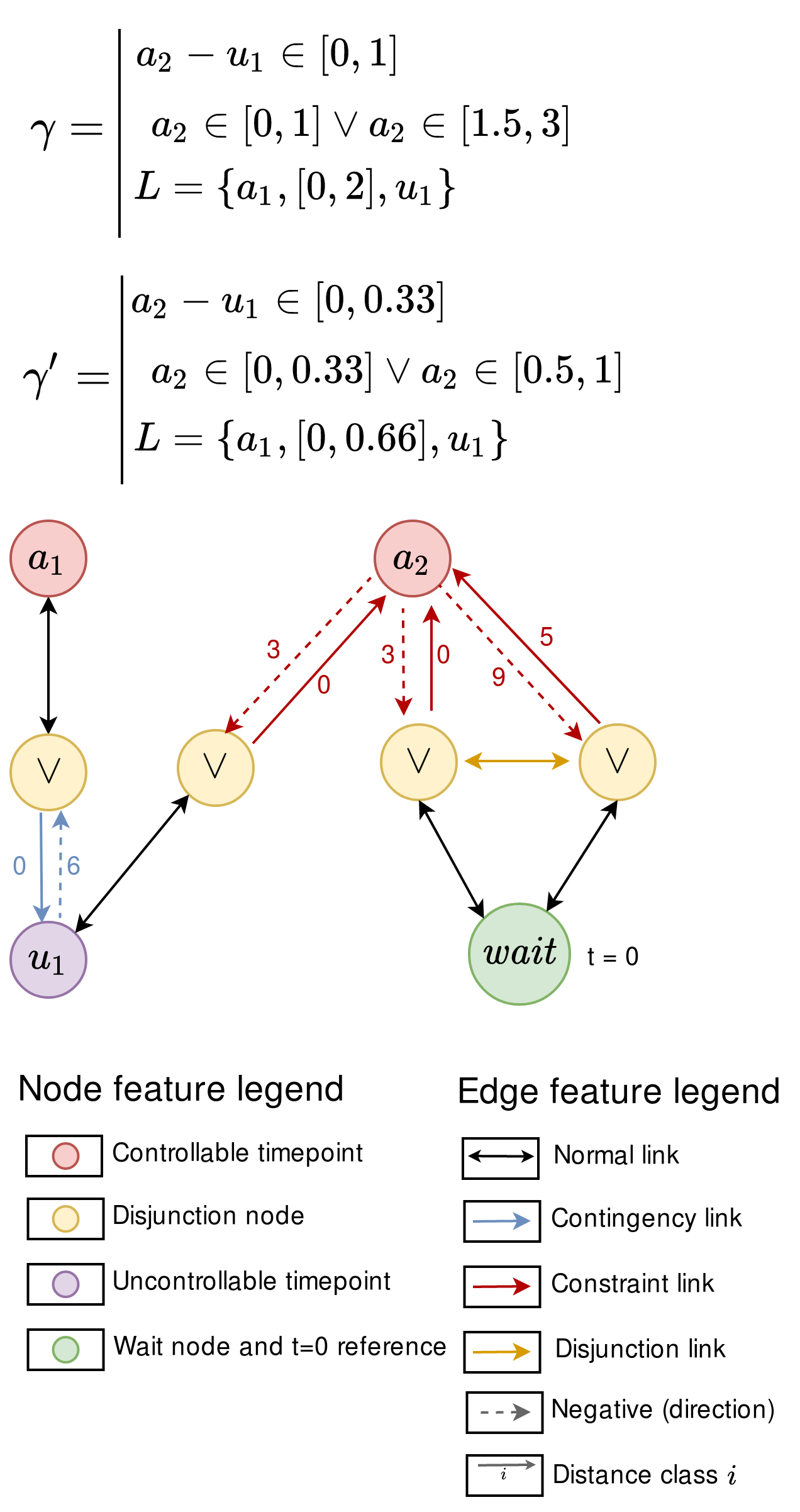}
\caption{\textbf{Conversion of a DTNU $\gamma$ into a graph.} $\gamma'$ is the normalized DTNU. Edge distances are expressed as distance classes. To distinguish between lower and upper bounds in intervals, we introduce an additional \textit{negative directional sign} feature.}
\label{fig-graph}
\end{figure}

Our MPNN architecture is made of 5 graph convolutional layers from \cite{gilmer2017neural} with 32 neurons in each layer. For each layer, we use a two-layer MLP (multi layer perceptron) with 128 neurons in the hidden layer to compute node and edge embeddings. In addition, we use batch normalization after each graph layer and apply the $\mathrm{ReLU}(\cdot) = \max(0,\cdot)$ activation function. The input of our MPNN is the graph conversion of a DTNU. Figure \ref{fig-graph} illustrates an example of graph conversion. We use $10$ different edge distance classes: $0:[0,0.1)$, $1:[0.1,0.2)$, ...,  $9:[0.9,1]$. Training is done with the \textit{adagrad} optimizer \cite{duchi2011adaptive} and an initial learning rate $10^{-4}$ on a dataset comprised of $30.000$ instances generated as described in \S \ref{randomized-tree-search}. We split the data into a training set comprised of $25.000$ instances and a cross-validation set comprised of $5.000$ instances. We add a dropout regularization layer with a \textit{keep rate} $0.9$ before the output layer to reduce overfitting.

\end{document}